\documentclass[10pt,twocolumn,letterpaper]{article}

\usepackage{cvpr}
\usepackage{times}
\usepackage{epsfig}
\usepackage{graphicx}
\usepackage{amsmath}
\usepackage{amssymb}
\newcommand{\cHiro}[1]{{\color{red}{#1}}}

\usepackage{color}
\usepackage{comment}

\usepackage[breaklinks=true,bookmarks=false]{hyperref}

\cvprfinalcopy 

\def\cvprPaperID{****} 

\begin{document}

\title{Neural Joking Machine:\\Humorous image captioning }

\author{{Kota Yoshida$^1$\thanks{denotes equal contribution}} , {Munetaka Minoguchi$^{1*}$}, Kenichiro Wani$^{1}$, Akio Nakamura$^1$ and Hirokatsu Kataoka$^2$\\
1) Tokyo Denki University\\
2) National Institute of Advanced Industrial Science and Technology (AIST)\\
{\tt\small {\{yoshida.k,minoguchi.m,wani.k\} @is.fr.dendai.ac.jp, 
nkmr-a@cck.dendai.ac.jp,}}\\
{\tt\small {hirokatsu.kataoka@aist.go.jp}}
}

\maketitle

\begin{abstract}
  What is an effective expression that draws laughter from human beings? In the present paper, in order to consider this question from an academic standpoint, we generate an image caption that draws a ``laugh" by a computer. A system that outputs funny captions based on the image caption proposed in the computer vision field  is constructed. Moreover, we also propose the \textit{Funny Score}, which flexibly gives weights according to an evaluation database. The Funny Score more effectively brings out  ``laughter" to optimize a model. In addition, we build a self-collected BoketeDB, which contains a theme (image) and funny caption (text) posted on ``Bokete", which is an image Ogiri\footnote{Ogiri is a simple game that involves providing funny answers to themes.} website. In an experiment, we use BoketeDB to verify the effectiveness of the proposed method by comparing the results obtained using the proposed method and those obtained using MS COCO Pre-trained CNN+LSTM, which is the baseline and idiot created by humans. We refer to the proposed method, which uses the BoketeDB pre-trained model, as the Neural Joking Machine (NJM). 

\end{abstract}


\section{Introduction}

Laughter is a special, higher-order function that only humans possess. In the analysis of laughter, as Wikipedia says, ``Laughter is thought to be a shift of composition (schema)", and laughter frequently occurs when there is a change from a composition of receiver.  However, the viewpoint of laughter differs greatly depending on the position of the receiver. Therefore, the quantitative measurement of laughter is very difficult. Image Ogiri on web services such as "Bokete"~\cite{bokete} have recently appeared, where users post funny captions for thematic images and the captions are evaluated in an SNS-like environment.  Users compete to obtain the greatest number of ``stars''. Although quantification of laughter is considered to be a very difficult task, the correspondence between evaluations and images on Bokete allows us to treat laughter quantitatively. Image captioning is an active topic in computer vision, and we believe that humorous image captioning can be realized. The main contributions of the present paper are as follows:

\begin{figure}
\begin{center}
  \includegraphics[width=1.0\linewidth]{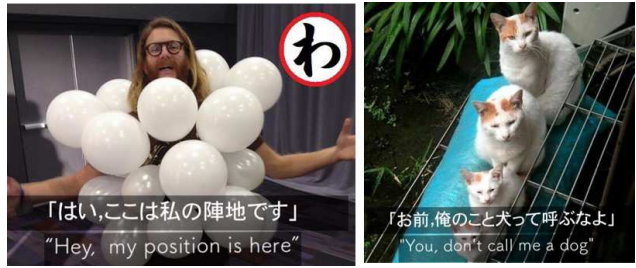}
\end{center}
  \caption{Examples of funny captions generated by NJM from an image input.}
\label{fig:lstm}
\end{figure}

\begin{figure*}
\begin{center}
   \includegraphics[width=0.85\linewidth]{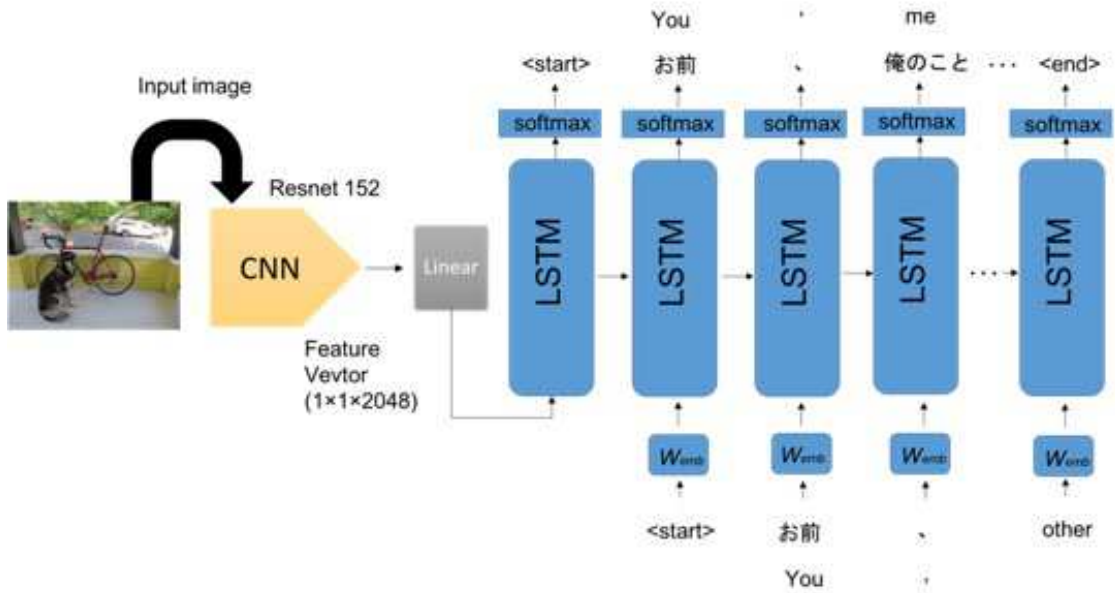}
\end{center}
   \caption{Proposed CNN+LSTM architecture for funny caption generation.}
\label{fig:lstm}
\end{figure*}


\begin{itemize}
\item We propose a framework for a funny caption generator based on recent image captioning work in the computer vision field. 
\item We define Funny Score, which is a weighting system based on the evaluation of existing funny captions in a database. The Funny Score is used in a loss function.
\item We have collected data to create BoketeDB from the web service Bokete. The database contains 999,571 image and caption pairs.
\end{itemize}BoketeDB

In the experimental section, we compare the proposed method based on Funny Score and BoketeDB pre-trained parameters with a baseline provided by MS COCO Pre-trained CNN+LSTM. We also compare the results of the NJM with funny captions provided by humans. In an evaluation by humans, the results provided by the proposed method were ranked lower than those provided by humans (22.59\% vs. 67.99\%) but were ranked higher than the baseline (9.41\%). Finally, we show the generated funny captions for several images.

\section{Related Research}
Through the great research progress with deep neural networks (DNNs), the combination of a convolutional neural network and a recurrent neural network (CNN+RNN) is a successful model for both feature extraction and sequential processing~\cite{LeCunNature2015}. 
Although there is no clear division, a CNN is often used for image processing, whereas an RNN is used for text processing. Moreover, these two domains are integrated. One successful application is image caption generation with CNN+LSTM (CNN+Long-Short Term Memory)~\cite{VinyalsCVPR2015}. This technique enables text to be automatically generated from an image input. However, we believe that image captions require human intuition and emotion. In the present paper, we help to guide an image caption has funny expression.  In the following, we introduce related research on humorous image caption generation.

Wang \textit{et al.} proposed an automatic ``meme" generation technique~\cite{WangNAACL2015}. A meme is a funny image that often includes humorous text. Wang~\textit{et al.} statistically analyzed the correlation between memes and comments in order to automatically generate a meme by modeling probabilistic dependencies, such as those of images and text.

Chandrasekaran \textit{et al.} conducted a humor enhancement of an image~\cite{ChandrasekaranCVPR2016} by constructing an analyzer to quantify ``visual humor'' in an image input. They also constructed datasets including interesting (3,200) and non-interesting (3,200) human-labeled images to evaluate visual humor. The ``funniness'' of an image can be trained by defining five stages. 


\section{Proposed Method}
We effectively train a funny caption generator by using the proposed Funny Score by weight evaluation. We adopt CNN+LSTM as a baseline, but we have been exploring an effective scoring function and database construction. We refer to the proposed method as the Neural Joking Machine (NJM), which is combined with the BoketeDB pre-trained model, as described in Section~\ref{sec:boketedb}.

\subsection{CNN+LSTM}

The flow of the proposed method is shown in Figure ~\ref{fig:lstm}. Basically, we adopted the CNN+LSTM model used in Show and Tell, but the CNN is replaced by ResNet-152 as an image feature extraction method. In the next subsection, we describe in detail how to calculate a loss function with a Funny Score. The function appropriately evaluates the number of stars and its ``funniness''.

\subsection{Funny Score}
The Bokete Ogiri website uses the number of stars\footnote{Something like ``good"  in an SNS.} to evaluate the degree of funniness of a caption. The user evaluates the ``funniness'' of a posted caption and assigns one to three stars to the caption. Therefore, funnier captions tend to be assigned a lot of stars. We focus on the number of stars in order to propose an effective training method, in which the Funny Score enables us to evaluate the funniness of a caption.
Based on the results of our pre-experiment, a Funny Score of 100 stars is treated as a threshold. In other words, the Funny Score outputs a loss value $L$ when \#star is less than 100. In contrast, the Funny Score returns $L - 1.0$ when \#star is over 100. The loss value $L$ is calculated with the LSTM as an average of each mini-batch.

\section{BoketeDB}
\label{sec:boketedb}
We have downloaded pairs of images and funny captions in order to construct a Bokete Database (BoketeDB). As of March 2018, 60M funny captions and 3.4M images have been posted on the Bokete Ogiri website. In the present study, we consider 999,571 funny captions for 70,981 images. A number of pair between image and funny caption is posted in temporal order on the Ogiri website Bokete.  We collected images and funny captions to make corresponding image and caption pairs. Thus, we obtained a database for generating funny captions like an image caption one.

\textbf{Comparison with MS COCO~\cite{MSCOCO}.}
MS COCO contains a correspondence for each of 160,000 images to one of five types of captions.  In comparison with MS COCO, BoketeDB has approximately half the number of the images and 124\% the number of captions.

\section{Experiment}
We conducted evaluations to confirm the effectiveness of the proposed method. We describe the experimental method in Section~\ref{sec:contents}, and the experimental results are presented in Section~\ref{sec:results}.

\begin{table}[t]
\begin{center}
\begin{tabular}{|l|crr|}
\hline
Caption method &First [\%]&Second [\%]&Third [\%] \\
\hline \hline
Human & \underline{\textbf{67.99}} &25.59  & 9.414 \\
NJM (Proposed)  & \textbf{22.59}   & 60.04 & 17.36  \\
STAIR caption~\cite{STAIR Caption}  & 9.414  & 17.36 & 73.22 \\
\hline
\end{tabular}
\caption{Comparison of the output results: The ``Human" row indicates captions provided by human users and was ranked highest on the Bokete website. The ``NJM" row indicates the results of applying the proposed model based of Funny Score and BoketeDB. The ``STAIR caption" row indicates the results provided by Japanese translation of MS COCO.}
\label{tab:data_type2}
\end{center}
\end{table}

\subsection{Experimental contents}
\label{sec:contents}
Here, we describe the experimental method used to validate the effectiveness of the NJM. We compare the proposed method with two other methods of generating funny captions: 1) human generated captions, which are highly ranked on Bokete (indicated by ``Human" in Table~\ref{tab:data_type2}), and 2) Japanese image caption generation using CNN+LSTM pre-trained by STAIR caption~\cite{STAIR caption}. Based on the captions provided by MS COCO, the STAIR caption is translated from English to Japanese (indicated by ``STAIR caption'' in Table~\ref{tab:data_type2}). We use a questionnaire as the evaluation method. We selected a total of 30 themes from the Bokete Ogiri website that included ``people'', ``two or more people'', ``animals'', ``landscape'', ``inorganics'', and ``illustrations''. The questionnaire asks respondents to rank the captions provided by humans, the NJM, and STAIR caption in order of ``funniness''. The questionnaire does not reveal the origins of the captions.

\subsection{Questionnaire Results}
\label{sec:results}
In this subsection, we present the experimental results along with a discussion. Table~\ref{tab:data_type2} shows the experimental results of the questionnaire. A total of 16 personal questionnaires were completed\footnote{The questionnaires were administered without any of the respondents being able to consult with each other.}. Table~\ref{tab:data_type2} shows the percentages of captions of each rank for each method of caption generation considered herein. Captions generated by humans were ranked ``funniest'' 67.99\% of the time, followed by the NJM at 22.59\%. The baseline captions, STAIR caption, were ranked ``funniest'' 9.41\% of the time. These results suggest that captions generated by the NJM are less funny than those generated by humans. However, the NJM is ranked much higher than STAIR caption.

\begin{figure*}[t]
\begin{center}
\includegraphics[width=1.0\linewidth]{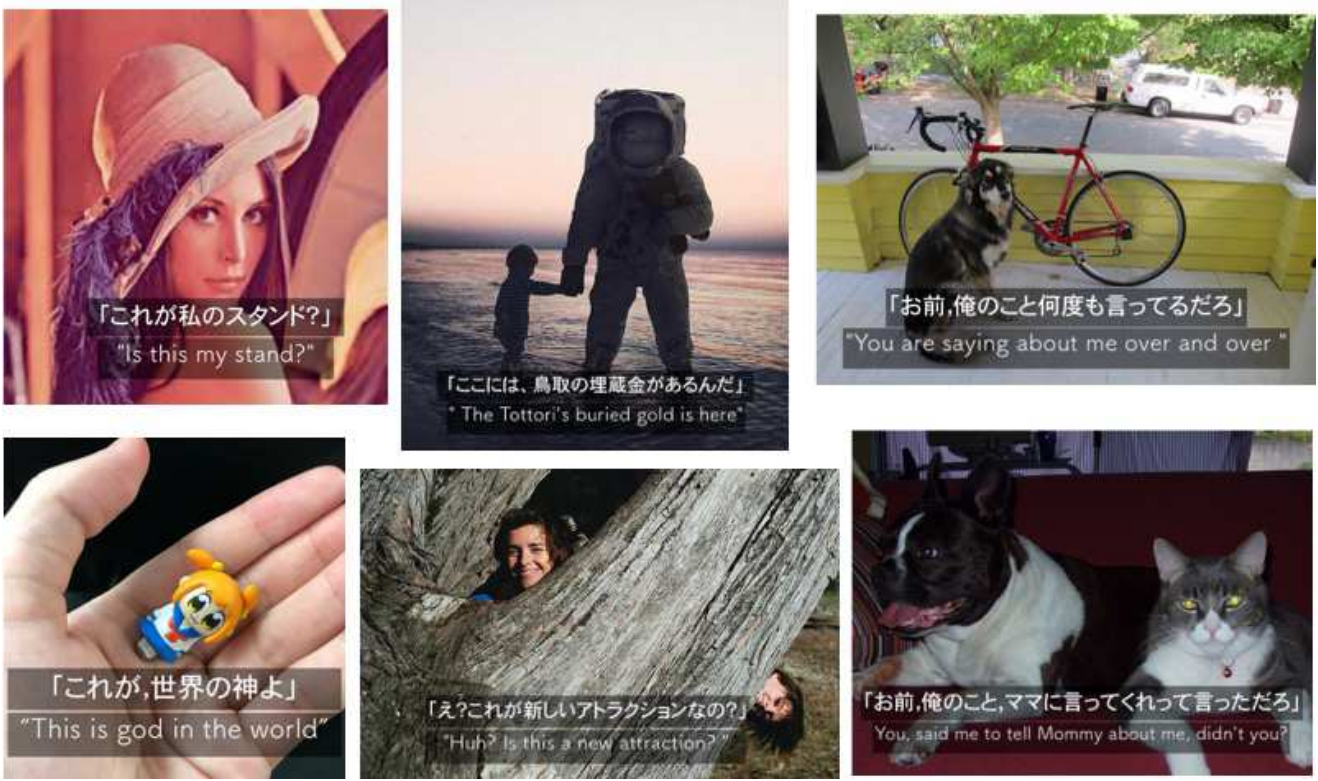}
\end{center}
    \caption{Visual results obtain using the proposed NJM.}\label{fig:iog4}
\end{figure*}

\subsection{Posting to Bokete}
We are currently posting funny captions generated by the NJM to the Bokete Ogiri website in order to evaluate the proposed method. Here, we compare the proposed method with STAIR captions. As reported by Bokete users, the funny captions generated by STAIR caption averaged 1.71 stars, whereas the NJM averaged \textbf{3.23} stars. Thus, the NJM is funnier than the baseline STAIR caption according to Bokete users. We believe that this difference is the result of using (i) Funny Score to effectively train the generator regarding funny captions and (ii) the self-collected BoketeDB, which is a large-scale database for funny captions. 

\subsection{Visual results}
Finally, we present the visual results in Figure~\ref{fig:iog4}, which includes examples of funny captions obtained using NJM. Although the original caption is in Japanese, we also translated the captions into English. Enjoy!


\section{Conclusion}
In the present paper, we proposed a method by which to generate captions that draw laughter. We built the BoketeDB, which contains pairs comprising a theme (image) and a corresponding funny caption (text) posted on the Bokete Ogiri website. We effectively trained a funny caption generator with the proposed Funny Score by weight evaluation. Although we adopted CNN+LSTM as a baseline, we have been exploring an effective scoring function and database construction. The experiments of the present study suggested that the NJM was much funnier than the baseline STAIR caption.


\begin{thebibliography}{9}
\bibitem{bokete}Omoroki.Inc, "Bokete," https://bokete.jp.
\bibitem{LeCunNature2015}
Yann LeCun, Yoshua Bengio, Geoffrey Hinton, 
"Deep Learning,"
{\it Nature 521 (7553):436-44}, 2015.

\bibitem{VinyalsCVPR2015}
Oriol Vinyals, Alexander Toshev, Samy Bengio, Dumitru Erhan,
"Show and Tell: A Neural Image Caption Generator,"
{\it In Proceedings of the IEEE Conference on Computer
Vision and Pattern Recognition (CVPR)}, 2015.

\bibitem{WangNAACL2015}
William Yang Wang and Miaomiao Wen,
``I Can Has Cheezburger? A Nonparanormal Approach to Combining Textual and Visual Information for Predicting and Generating Popular Meme Descriptions,"
{\it In proceedings of the North American Chapter of the Association for Computational Linguistics(NAACL)}, 355-365, Denver, Colorado, May 31 - June 5, 2015.

\bibitem{ChandrasekaranCVPR2016}
Arjun Chandrasekaran, Ashwin K. Vijayakumar, Stanislaw Antol, Mohit Bansal,
Dhruv Batra, C. Lawrence Zitnick and Devi Parikh, 
"We Are Humor Beings: Understanding and Predicting Visual Humor,"
{\it In Proceedings of the IEEE Conference on Computer
Vision and Pattern Recognition (CVPR)}, 2016.

\bibitem{MSCOCO}
Tsung-Yi Lin, Michael Maire, Serge Belongie, James Hays, Pietro Perona, Deva Ramanan, Piotr Dollár, C. Lawrence Zitnick, ``Microsoft COCO: Common Objects in Context," {\it in ECCV}, 2014.

\bibitem{STAIR Caption}
Yuya Yoshikawa, Yutaro Shigeto and Akikazu Takeuchi, "STAIR Captions: Constructing a Large-Scale Japanese Image Caption Dataset,"
{\it In proceedings of Annual Meeting of the ACL - Association of Computational Linguistics(ACL)}, 2017.
\end{thebibliography}

{\small

\end{document}